\documentclass[10pt, conference]{IEEEtran}
\IEEEoverridecommandlockouts
\usepackage{cite}
\usepackage{amsmath,amssymb,amsfonts}
\usepackage{algorithmic}
\usepackage{graphicx}
\usepackage{textcomp}
\usepackage{xcolor}
\usepackage{svg}
\usepackage{subfig}
\usepackage{booktabs} 
\usepackage{multirow} 
\usepackage{pifont}
\usepackage{colortbl}
\usepackage{comment}

\definecolor{darkgreen}{rgb}{0.0, 0.5, 0.0}
\definecolor{darkred}{rgb}{0.8, 0.0, 0.0}
\definecolor{darkblue}{rgb}{0.0, 0.0, 0.8}

\usepackage[a4paper, total={184mm,239mm}]{geometry}
\def\BibTeX{{\rm B\kern-.05em{\sc i\kern-.025em b}\kern-.08em
    T\kern-.1667em\lower.7ex\hbox{E}\kern-.125emX}}

\newcommand{\ml}[1]{{\color{red}\bf [ML: #1]}}
    
\begin{document}

\title{LightMamba: Efficient Mamba Acceleration on FPGA with Quantization and Hardware Co-design}

\author{
    Renjie Wei$^{12*}$,
    Songqiang Xu$^{5*}$, Linfeng Zhong$^{6*}$, Zebin Yang$^{12}$, Qingyu Guo$^2$,\\
    Yuan Wang$^{23}$, Runsheng Wang$^{234}$ and Meng Li$^{123\dag}$
\\
\textit{$^1$Institute for Artificial Intelligence \& $^2$School of Integrated Circuits, Peking University, Beijing, China} \\
\textit{$^3$Beijing Advanced Innovation Center for Integrated Circuits, Beijing, China} \\
\textit{$^4$Institute of Electronic Design Automation, Peking University, Wuxi, China} \\
\textit{$^5$School of Software and Microelectronics, Peking University, Beijing, China} \\ 
\textit{$^6$School of Electronic and Computer Engineering, Peking University, Shenzhen, China}

\thanks{
This work was supported in part by National Natural Science Foundation of China under Grant 62495102 and Grant 92464104, in part by Beijing Municipal Science and Technology Program under Grant Z241100004224015, and in part by 111 Project under Grant B18001. 

$^*$Equal contribution.
$^\dag$Corresponding author.}
}

\maketitle

\begin{abstract}

State space models (SSMs) like Mamba have recently attracted much attention.
Compared to Transformer-based large language models (LLMs), Mamba achieves
linear computation complexity with the sequence length and demonstrates superior performance.
However, Mamba is hard to accelerate due to the scattered activation outliers and the complex computation dependency, rendering existing LLM accelerators inefficient.
In this paper, we propose LightMamba  that co-designs the quantization algorithm
and FPGA accelerator architecture for efficient Mamba inference. 
We first propose an FPGA-friendly post-training quantization algorithm that features
rotation-assisted quantization and power-of-two SSM quantization to reduce the majority of computation
to 4-bit. 
We further design an FPGA accelerator that partially unrolls the Mamba computation
to balance the efficiency and hardware costs. 
Through computation reordering as well as fine-grained
tiling and fusion, the hardware utilization and memory efficiency of the accelerator get drastically
improved. 
We implement LightMamba on Xilinx Versal VCK190 FPGA
and achieve 4.65$\sim$6.06$\times$ higher energy efficiency over the GPU baseline.
When evaluated on Alveo U280 FPGA, LightMamba reaches 93 tokens/s, 
which is 1.43$\times$ that of the GPU baseline.
Our code is available at https://github.com/PKU-SEC-Lab/LightMamba.



\end{abstract}

\section{Introduction}
\label{sec:introduction}

State space models (SSMs) like Mamba \cite{gu2023mamba,dao2024transformers} have recently been proposed
and emerged as a promising class of architectures as foundation models. Compared to existing Transformer-based
large language models (LLMs) \cite{touvron2023llama,touvron2023llama2,bubeck2023sparks,jiang2024mixtral},
Mamba only requires linear computational complexity with the increase of input sequence length while demonstrating
superior performance on various downstream tasks. 

\begin{figure}[!tb]
\centering
\includegraphics[width=\linewidth]{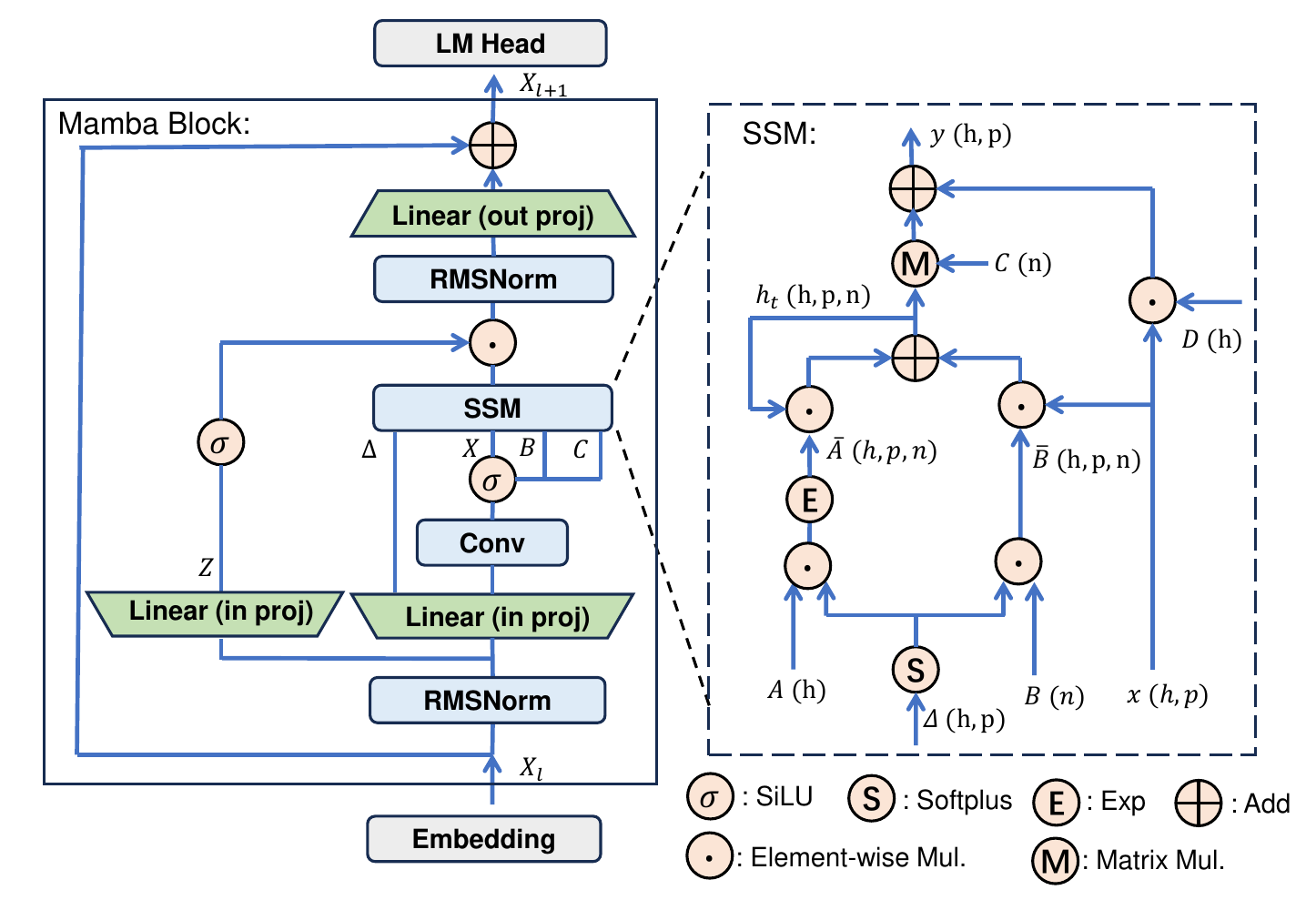}
\caption{The model architecture of Mamba2 and the detailed
computation graph of the SSM layer.} 
\vspace{-10pt}
\label{fig:model_arch}
\end{figure}

The basic architecture 
of Mamba~\cite{dao2024transformers} is shown in Fig.~\ref{fig:model_arch}. Each Mamba block mainly consists of two linear projection layers,
i.e., input projection and output projection, a 1-dimensional convolution (conv1d) layer, and an SSM layer. The computation of
Mamba involves a prefill stage that summarizes the input prompts and an autoregressive decode stage to produce the output tokens.
The decode of Mamba only generates and stores a fixed-size hidden state instead of a key-value cache in Transformers
that grows linearly with the sequence length. While Mamba has different architectures, we focus on the
Mamba2 \cite{dao2024transformers} architecture and all subsequent references to Mamba refer to Mamba2 unless otherwise specified.

Though promising, how to accelerate Mamba processing, especially on reconfigurable platforms like FPGA, remains
an open question. We observe existing works on FPGA-based LLM acceleration \cite{Flightllm,DFX} cannot be directly applied
due to the following challenges. 
First, while FPGA-based acceleration benefits from low-bit-precision quantization,
Mamba quantization is more challenging than Transformer-based LLMs due to scattered activation outliers. 
Specifically, for different input tokens, the activation outliers appear in different channels.
Naively applying existing quantization algorithms \cite{xiao2023smoothquant,wei2022outlier,wei2023outlier} incurs significant accuracy degradation.
Second, an SSM layer involves excessive element-wise computation.
While existing works \cite{li2024evaluating,pierro2024mamba,li2024marca} often ignore the quantization of SSM layers
and suffer from a high hardware cost, directly quantizing the SSM layer also introduces significant re-quantization
overhead. Third, the SSM layer involves complex computation and data dependency, which prevents unfolding the SSM
computation spatially to improve the acceleration throughput.

To solve these challenges, we propose LightMamba, the first FPGA-based Mamba acceleration framework that co-designs
the quantization algorithm and accelerator architecture. 
We propose a rotation-assisted post-training quantization (PTQ)
algorithm for Mamba to remove the scattered outliers, which enables us to quantize the model to 4-bit with minimum 
accuracy degradation and significantly improve the communication efficiency with off-chip memory. 
For the SSM layer,
we propose a FPGA-friendly power-of-two (PoT) quantization scheme to realize re-quantization with simple shifting for better
computation efficiency. 
For the FPGA accelerator architecture, we design customized rotation modules for the PTQ algorithm
and further propose computation reordering as well as fine-grained tiling and fusion to improve the hardware utilization.
Our main contributions can be summarized as follows:
\begin{itemize}
    \item We propose the first PTQ algorithm for the entire Mamba model.
    Through rotation-assisted quantization, we quantize Mamba to 4-bit with plausible accuracy.
    The SSM layer is also quantized with FPGA-friendly PoT scheme for better computation efficiency.
    \item We propose the first FPGA-based Mamba accelerator. The architecture features a customized rotation module
    co-designed with the proposed quantization algorithm. We also propose computation reordering and
    fine-grained scheduling to improve the computation throughput and reduce the on-chip memory usage.
    \item We implement LightMamba on Xilinx Versal VCK190 FPGA achieving 7.21 tokens/s and 4.65$\sim$6.06$\times$ higher energy efficiency over the GPU baseline. 
\end{itemize}

\section{related works}
\label{sec:background}



\subsection{LLM Quantization}
Quantization maps the weights and activations from high-bit-precision floating point (FP) numbers,
e.g., FP16 to low-bit integers, e.g., INT8 or INT4,
reducing both memory and computation costs.
Existing algorithms mostly leverage PTQ for LLMs and observe
the key obstacle comes from the outliers in weights and activations.
LLM.int8()~\cite{dettmers2022llm} uses mix-precision quantization and
keeps outliers in high bit-precision,
which introduces large computation overhead.
SmoothQuant~\cite{xiao2023smoothquant} and 
Outlier Suppression~\cite{wei2022outlier,wei2023outlier} observe the outliers persist
in specific channels and thus, calculate 
the channel-wise scaling or shifting factors to rescale the outliers before quantization.
QuaRot~\cite{ashkboos2024quarot} and SpinQuant~\cite{liu2024spinquant}
multiply the weight and activation with a rotation matrix 
to remove outliers.
However, these methods are only demonstrated for Transformer-based LLMs.
For Mamba, \cite{li2024evaluating} only quantizes the linear layers with round-to-nearest (RTN) method. 
Mamba-PTQ~\cite{pierro2024mamba} simply applies SmoothQuant on Mamba.
Although they only quantize the linear layers and leave the SSM layer in FP, they still
suffer from large accuracy degradation.

\subsection{FPGA-based LLM Accelerators}


Previous FPGA-based LLM accelerators can be categorized into two types: temporal architecture and spatial architecture. 
Temporal architecture constructs a processing engine (PE) performing various tasks, especially for matrix multiplication (MM)~\cite{DFX,Flightllm,enhancing}. 
This architecture is not well-suited for handling the diverse and complex element-wise multiplications (EM) in Mamba.
Spatial architecture customizes PEs for different operations and supports concurrent processing of multiple PEs 
in pipeline~\cite{understanding,HGPIPE,ftrans,adaptable},
leading to low latency.
However, due to the custom PEs dispersing resources, this design results in lower parallelism for MM.
In this paper, we adopt a partially unfolded spatial architecture that unfolds 
one Mamba block and co-designs the quantization algorithm and architecture to improve hardware utilization and
reduce the latency.
A comparative analysis of these architectures is presented in Table~\ref{tab:Qualitative comparison}.

\begin{table}[!t]
\centering
\caption{Qualitative comparison between different paradigms. W4A4 indicates 4-bit weight and 4-bit activation.}
\label{tab:Qualitative comparison}
\begin{tabular}{c|c|c|c}
\toprule
 & \cite{understanding} & \cite{Flightllm}\cite{DFX} & Ours\\
\midrule
\textbf{Architecture} & Spatial & Temporal & Partial Spat. \\
\textbf{Model} & \textcolor{black}{Transformer} & \textcolor{black}{Transformer} & \textcolor{black}{Mamba}\\
\textbf{Bit Precision} & W4A8 & W3.5A8 or FP16 & W4A4 \\
\textbf{Latency} & \textcolor{darkgreen}{Low} & \textcolor{darkred}{High} & \textcolor{darkgreen}{Low}\\
\textbf{EM Compatibility} & \textcolor{darkgreen}{$\checkmark$} & \textcolor{darkred}{$\times$} & \textcolor{darkgreen}{$\checkmark$}\\
\textbf{MM parallelism} & \textcolor{darkblue}{Mid} & \textcolor{darkgreen}{High} & \textcolor{darkgreen}{High}\\

\bottomrule
\end{tabular}
\end{table}

\section{Motivation}
\label{sec:motivation}

\begin{table}[!tb]
\centering
\caption{4-bit quantization error of the activation in the out project layer in Mamba2-2.7B
with different PTQ methods. }
\label{tab:quant_error}
\begin{tabular}{c|c|c|c|c}
\toprule
Method       & RTN  & SQ~\cite{xiao2023smoothquant}   & OS+~\cite{wei2023outlier}   & Ours \\ \midrule
Quantization Error & 19.5 & 18.8 & 309.8 & 13.1 \\ \bottomrule
\end{tabular}
\end{table}



  
  

  

\begin{figure}[!tb]
    \centering
    \includegraphics[width=0.8\columnwidth]{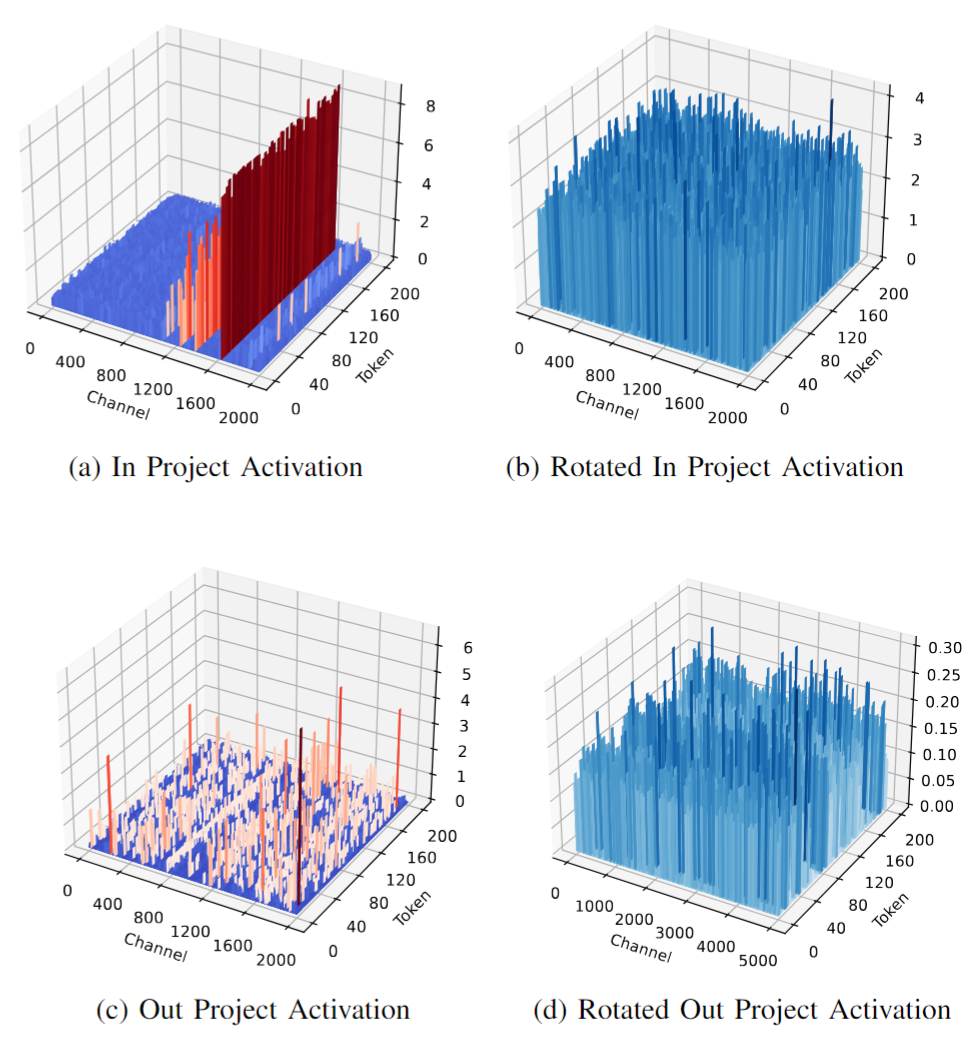}
    \caption{Activation distribution in Mamba2-2.7B before and after rotation.}
    \vspace{-10pt}
    \label{fig:activation_distribution}
\end{figure}


\textbf{Challenge 1:}
Mamba is difficult to quantize to low bit precision because of the scattered outliers.
We find that the activation outliers of the output projection layer exhibit scattered distribution
as shown in Fig.~\ref{fig:activation_distribution}(c),
which renders previous LLM quantization~\cite{xiao2023smoothquant,wei2022outlier,wei2023outlier} less effective.
This is because in Transformer-based LLMs, outliers persist in fixed channels across different tokens.
Hence, it is possible to calculate channel-wise scaling or shifting factors to reduce the magnitude of outlier channels and reduce the quantization error \cite{xiao2023smoothquant}.
For Mamba, activation outliers may show up in different channels for different input tokens and may scatter across all channels.
As a result, prior art methods, e.g., SmoothQuant \cite{xiao2023smoothquant} and OutlierSuppression+ \cite{wei2022outlier}, have comparable or even larger quantization error compared to the baseline RTN quantization as in Table~\ref{tab:quant_error}.
Rotation-based quantization algorithms have also been demonstrated for Transformer-based LLMs \cite{ashkboos2024quarot,liu2024spinquant} and do not require outliers to persist in fixed channels. 
However, as rotation-based algorithms usually require operator fusion and introduce extra computation overhead, it is unclear how it can be applied to Mamba acceleration on FPGAs.

\begin{figure}[!tb]
\centering
\includegraphics[width=0.9\linewidth]{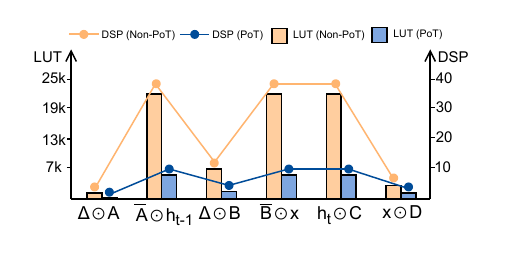}
\vspace{-10pt}
\caption{The hardware cost of different operations in the SSM layer with naive Non-PoT quantization and PoT quantization.} 
\vspace{-10pt}
\label{fig:challenge2}
\end{figure}

\textbf{Challenge 2:}
Existing works~\cite{li2024evaluating,pierro2024mamba,li2024marca} often choose not to quantize the SSM layer in Mamba. 
The floating-point (FP) computation of SSM layers introduces large hardware cost on FPGA.
Moreover, directly quantizing SSM layers to lower bit precision also incurs large hardware cost as shown in Fig.~\ref{fig:challenge2}. 
This is because the SSM layer comprises excessive EMs, the re-quantization (transforming output from high bit precision back to low bit precision) overhead of which is significantly large.

\textbf{Challenge 3:}
The computation graph of SSM layer is complex, which increases the difficulty of FPGA-based
acceleration. On one hand, the inputs of the SSM layer, i.e., $X, B, C, \Delta$, are all
generated by the input projection layer. 
Even though it is possible to unroll the Mamba computation
spatially on FPGA, the computation of the input projection and SSM layers are forced to be sequential,
leading to only less than 60\% hardware utilization. 
On the other hand, storing the intermediate activations of SSM on chip is memory-consuming, accounting for more than 70\% of the total URAM usage.



To address these challenges, 
we propose LightMamba, 
an efficient FPGA-based Mamba acceleration framework.
Specifically, we propose an FPGA-friendly Mamba quantization algorithm in Sec.~\ref{sec:quantization_algo},
featuring a rotation-assisted PTQ algorithm to mitigate the scattered outliers
and a PoT-based SSM quantization with minimum re-quantization overhead.
We also propose an FPGA-based accelerator in Sec.~\ref{sec:Hardware design},
featuring computation reordering to improve the  hardware utilization
and fine-grained tiling and fusion to reduce the on-chip memory cost.

\section{FPGA-friendly Quantization Algorithm}
\label{sec:quantization_algo}


\subsection{Rotation-assisted Linear Layer Quantization}

\begin{figure}[!tb]
  \centering 
  \subfloat[]{
    \label{subfig: quantization_algorithm}
    \includegraphics[width=0.25\textwidth]{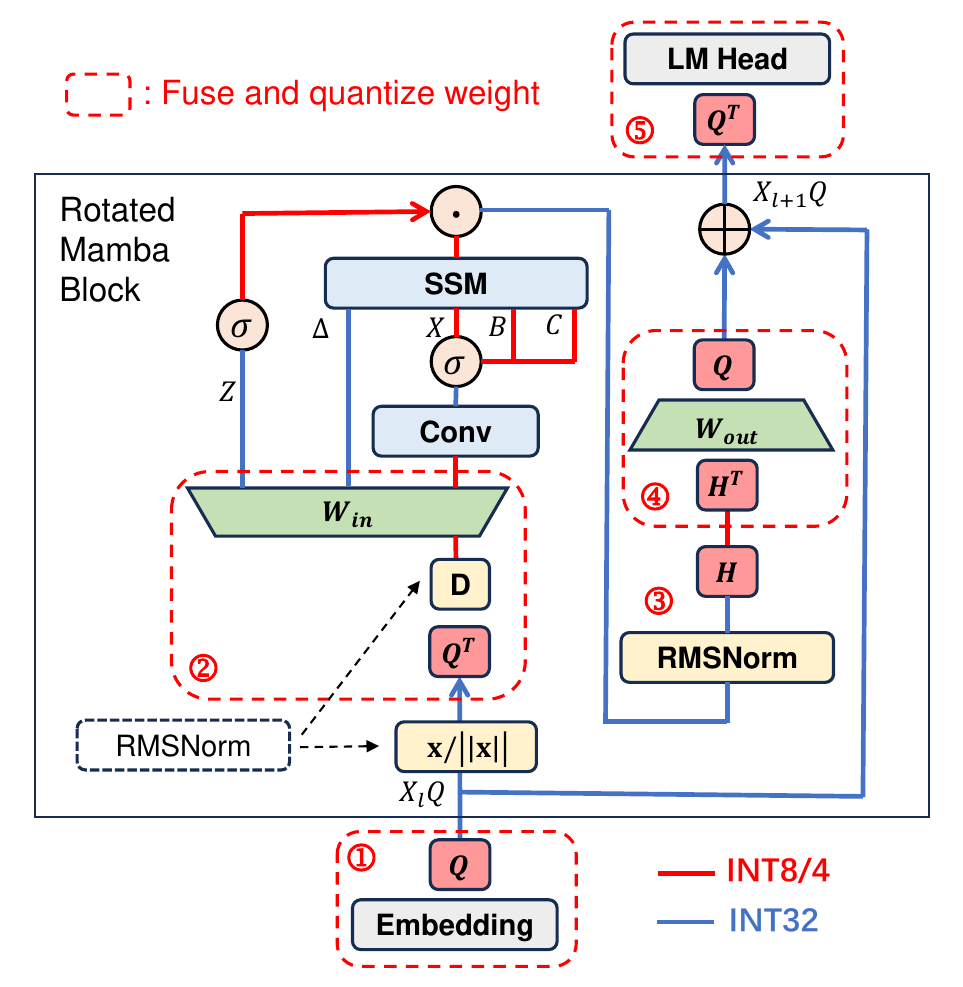}
  }
  \subfloat[]{
    \label{subfig: weight_quant_error}
    \includegraphics[width=0.23\textwidth]{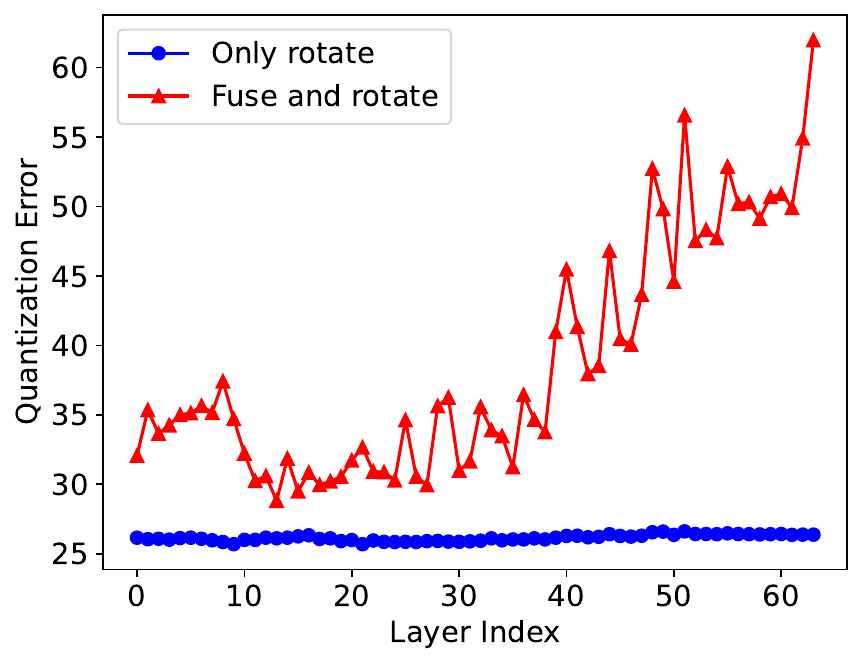}
  }
  \caption{
(a) The proposed rotation-assisted quantization algorithm.
Both $Q$ and $H$ are Hadamard matrices to ensure computation correctness.
(b) Quantization error of the output projection weight after only rotation or fusion and rotation.
}
  \label{fig:fusion_cause_error}
\end{figure}

The rotation-assisted quantization method is first proposed in~\cite{ashkboos2024quarot}
for Transformer-based LLMs.
By multiplying the activation $X$ and weight $W$ with orthogonal matrix $Q$, i.e., $XQQ^TW$,
the result is identical with $XW$,
while the outliers in $X$ and $W$ are removed.
However, it is still unclear whether the rotation method is applicable to Mamba.
Therefore, we study the rotation equivalence in Mamba and propose a rotation-assisted method shown in Fig.~\ref{subfig: quantization_algorithm}.


\begin{figure*}[!tb]
    \centering
    \includegraphics[width=1\linewidth]{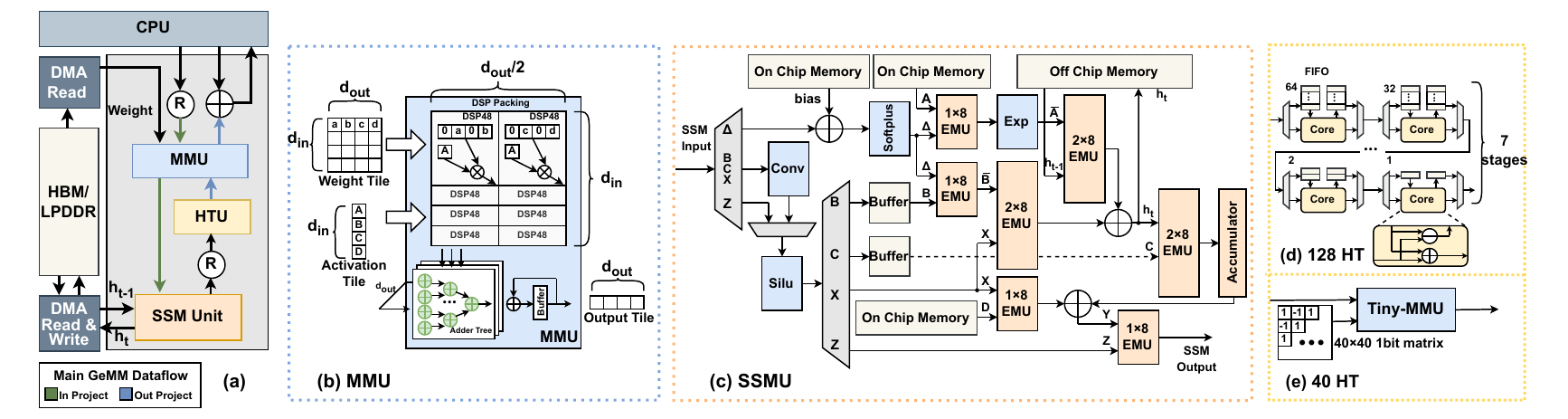}
    \vspace{-10pt}
    \caption{Diagram of (a) the overall architecture, (b) SSMU, (c) MMU, (d) 128-point HTU, and (e) 40-point HTU.}
    \vspace{-10pt}
    \label{fig: Hardware Design}
\end{figure*}

We observe that the activations in the linear layers and SSM layer
have large number of outliers,
and the outliers of output projection layer exhibit scattered distribution
across different channels.
Rotation is helpful to remove outliers
since it amortizes large outliers with other elements.
For the input and output projection layers,
we apply rotation and remove the outliers as shown in Fig.~\ref{fig:activation_distribution}.
It is worth noting that
to rotate the activation before the output projection layer
we insert an on-line Hadamard transformation 
before it in Fig.~\ref{subfig: quantization_algorithm},
which can be efficiently performed by our customized
rotation unit in Sec.~\ref{sec:Hardware design}.
However, we find that SSM cannot be rotated since it
does not satisfy the rotation equivalence.
Specifically,
the original computation in SSM is Eq.~\ref{eq:rotate_ssm_a}.
Assuming we can rotate hidden state $h_t$ to remove the outliers,
i.e., multiply $h_t$ by Hadamard matrix $H$,
we can derive Eq.~\ref{eq:rotate_ssm_b} and Eq.~\ref{eq:rotate_ssm_c}.
However, Eq.~\ref{eq:rotate_ssm_c} cannot derive Eq.~\ref{eq:rotate_ssm_d}
because EM does not satisfy matrix associative property.
Thus we cannot derive Eq.~\ref{eq:rotate_ssm_d} from Eq.~\ref{eq:rotate_ssm_a}, i.e., SSM does not satisfy the rotation equivalence.
\begin{subequations}
\begin{align}
    & h_{t} = \bar{A} \odot h_{t-1}+\bar{B} \odot X_{t}  \label{eq:rotate_ssm_a}\\
    & h_{t}H = (\bar{A} \odot h_{t-1} + \bar{B} \odot X_{t})H \label{eq:rotate_ssm_b}\\
    & h_{t}H = (\bar{A} \odot h_{t-1})H + (\bar{B} \odot X_{t})H \label{eq:rotate_ssm_c}\\
    & h_{t}H = \bar{A} \odot (h_{t-1}H) + \bar{B} \odot (X_{t}H) \label{eq:rotate_ssm_d}
\end{align}
\end{subequations}

To reduce the computation overhead, we try to fuse rotation with neighboring operations as much
as possible. 
As shown in Fig.~\ref{subfig: quantization_algorithm}, we can fuse the first rotation
with the embedding table (i.e., \textcircled{1}),
the last rotation with the LM head (i.e., \textcircled{5}),
as well as the rotations before and after the output projection layers in each Mamba block (i.e., \textcircled{4}).
For the rotation next to the first RMSNorm operator (i.e., \textcircled{2}), to ensure
the computational invariance, we need to split the scaling factor, i.e., $D$, of the RMSNorm first,
and then, fuse it with the weights of input projection.
For the rotation next to the second RMSNorm operator (i.e., \textcircled{3}), we find whether
or not to fuse the scaling factor of the RMSNorm to the weight of output projection does not impact the computational invariance,
while fusion introduces a larger quantization error as in Fig.~\ref{subfig: weight_quant_error}. Hence, we choose not to fuse
the scaling factor of the second RMSNorm.
In our algorithm, only rotation \textcircled{3} needs to be computed online, which incurs small computation
overhead with our customized FPGA module support.

%

\subsection{FPGA-friendly SSM Quantization}

In order to quantize SSM to reduce the heavy hardware cost by FP computations,
we leverage INT8 per-group quantization
to strike a balance between accuracy and hardware efficiency.
However, directly quantizing SSM introduces
large re-quantization overhead as shown in Fig.~\ref{fig:challenge2},
which is
because EM has larger re-quantization overhead than MM intrinsically since there is no reduction in EM.
To this end,
we propose to use PoT quantization for SSM,
through which re-quantization can be implemented in bit-shifting
rather than multiplication
thus reducing the re-quantization overhead significantly.

\section{FPGA-based Mamba Acceleration}
\label{sec:Hardware design}



\subsection{Overall Hardware and Hadamard Transform Unit}
\label{subsec:Overall_architecture}



The overall architecture of LightMamba is shown in Fig.~\ref{fig: Hardware Design} (a). 
We design a partially unfolded spatial architecture and unroll the computation of one Mamba block on the FPGA.
The model parameters are stored in the off-chip DRAM.
LightMamba consists of three main modules: (1) Matrix Multiplication Unit (MMU), which handles the computations of all linear layers, including both input and output projection; (2) SSM Unit (SSMU), which fully unfolds SSM to enable pipelined execution; and (3) Hadamard Transform Unit (HTU), a customized design to support rotation-assisted quantization.

\textbf{MMU}
is designed to support input and output projection layers in a time-multiplexed manner.
It features a tree-based architecture of multiplier-accumulators (MACs) that receive vectors of $d_{in}$ dimension as inputs. 
It is also equipped with $d_{out}$ lanes, which performs $d_{in} \times d_{out}$ MACs in one cycle. Altogether, MMU contains $d_{in} \times d_{out}$ MAC units, which are efficiently implemented using ${d_{in} \times d_{out}}/2$ DSPs, leveraging the DSP packing technique~\cite{pack}, as illustrated in Fig.~\ref{fig: Hardware Design}(b).




\textbf{SSMU}
features a fine-grained, fully pipelined architecture 
for computing the SSM layer.
As shown in Fig.~\ref{fig: Hardware Design}(c), 
each operator is implemented by a dedicated unit, connected via first-in-first-out buffers (FIFOs). 
We optimize the parallelism for each operator to ensure a balanced data flow with a minimum FIFO depth.
We implement the operators in SSM through Element-wise Multiplication Units (EMUs) which are composed of DSPs. 
We set different parallelism for different operators. 

\textbf{HTU}
is dedicated to support the Hadamard transformation in our quantization algorithm. 
It has two variants: the power-of-two and the non-power-of-two type. 
For example, in Mamba-2.7B,  two types of HTU are required,
i.e, 128-point HTU and 40-point HTU as in Fig.~\ref{fig: Hardware Design}(d) and Fig.~\ref{fig: Hardware Design}(e).
The 128-point HTU is based on the Fast Hadamard Transformation (FHT) algorithm~\cite{FHT}.
It contains seven stages, each containing a Butterfly Core and two FIFOs.
In the first stage, the first 64 elements are pushed into the input FIFO.
When the next 64 elements arrive, they are processed in pairs by the Butterfly Core. 
Outputs are either sent to the next stage or buffered in the output FIFO. 
Compared to the MM-based Hadamard transform, this design reduces latency by 72\% with the same hardware resources.
For the small 40-point Hadamard transformation, we directly implement it with a simple MMU and fix one input to the Hadamard matrix with only 1 and -1.



\begin{figure}[!tb]
    \centering
    \includegraphics[width=1\columnwidth]{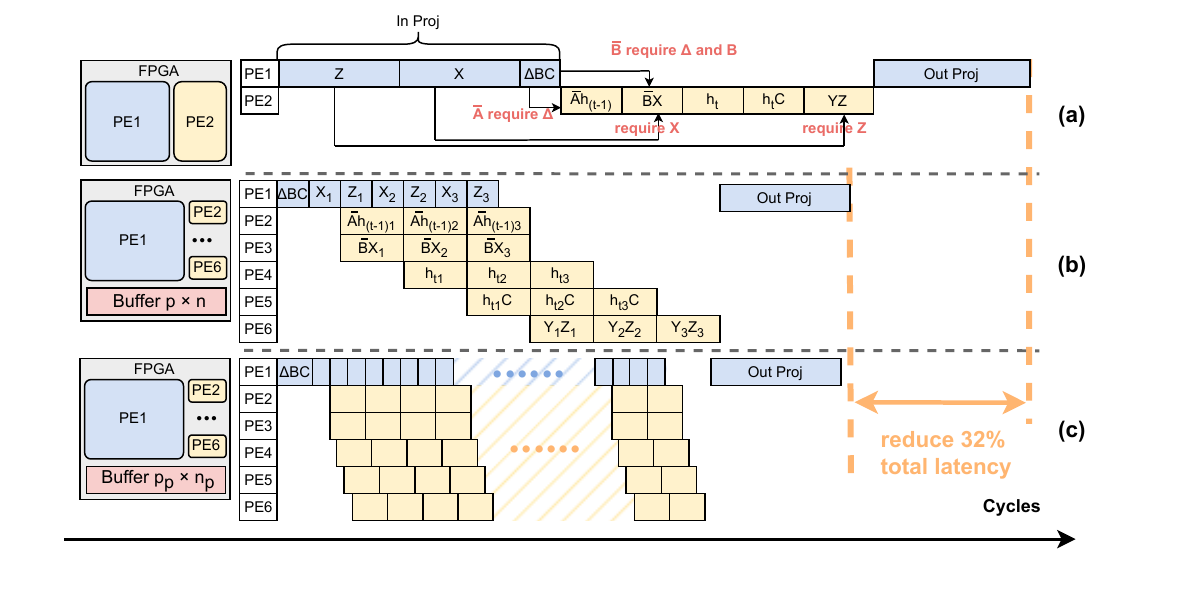}
    \vspace{-20pt}
    \caption{Pipeline scheme: (a) Naive implementation, (b) Coarse-grained pipeline, (c) Fine-grained pipeline.}
    \label{fig: Pipeline scheme}
\end{figure}

\subsection{Computation Reordering}
\label{subsec:pipeline}

Due to the distinct computational patterns
and the data dependency between SSM and input projection layer,
they are forced to execute sequentially as in Fig.~\ref{fig: Pipeline scheme}(a).
However, we find that SSM comprises multiple independent heads, and propose a coarse-grained pipeline to improve hardware utilization as in Fig.~\ref{fig: Pipeline scheme}(b). 
This pipeline design depends on our proposed computation reordering method, which alters the data generation sequence in the input projection layer. 
Specifically, $\Delta, B, C$ are generated first and stored in an on-chip buffer, while $X$ and $Z$ are produced alternatively for computing SSM head-by-head, as shown in Figure~\ref{fig: Pipeline scheme}(b).
This reordering allows the SSM computation to begin immediately after $\Delta, B, C$ are produced in the MMU. Compared to the traditional sequential implementation, our approach reduces the total computation time of the network by 32\% and increases hardware utilization from 58\% to 96\%, as depicted in Figure~\ref{fig: Pipeline scheme}(b).

\subsection{Fine-grained Tiling and Fusion}
\label{subsec: tiling and fusion}
\begin{figure}[!tb]
    \centering
    \includegraphics[width=0.8\columnwidth]{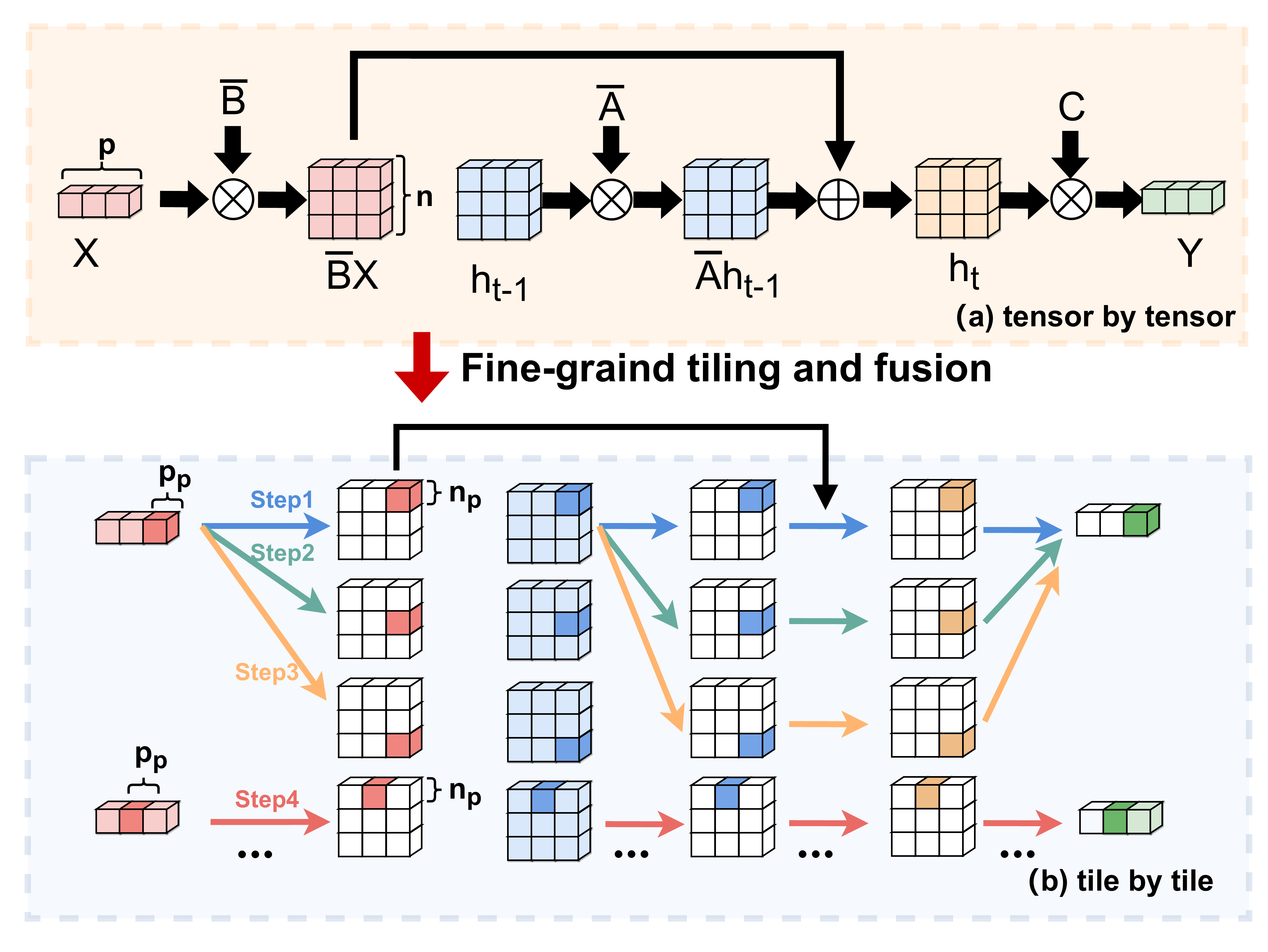}
    \caption{Fine-grained tiling and fusion.}
    \vspace{-10pt}
    \label{fig:tiling}
\end{figure}

Given the current design, we observe the SSMU consumes more than 70\% of the on-chip memory as it requires to store all intermediate activations, 
e.g., $\Bar{B}X$, $\Bar{A}{h_{t-1}}$, and $h_t$, etc, as in Fig.\ref{fig:tiling}(a).
We propose a fine-grained tiling and fusion strategy. 
By leveraging operation fusion, we directly feed the output of the previous operator to the next operator,
and thus eliminating the on-chip communication and data handling.
Additionally, fine-grained tiling is employed to reduce the buffer size.
We tile along the head and hidden state dimensions, with a tile size of $n_p \times p_p$ as in Fig.~\ref{fig:tiling}(b).
The tile-by-tile implementation refines the execution of SSMU to enable the fine-grained pipeline in Fig.~\ref{fig: Pipeline scheme}(c), 
which reduces the URAM usage of SSMU by 4$\times$. 
This not only lower the on-chip buffer requirements, but also eliminates the pipeline bubbles for better hardware utilization.

\section{experiments}
\label{sec:experiments}

\subsection{Experiment Setup}
\label{subsec:Evaluation Setup}

\textbf{Algorithm}
We evaluate our proposed quantization algorithm
on the Mamba model family~\cite{dao2024transformers}.
We quantize the entire model shown in Fig.~\ref{subfig: quantization_algorithm}
for hardware efficiency.
We use per-channel weight quantization 
and per-token activation quantization for 8-bit weight and activation quantization (denoted as W8A8).
We use per-group weight and activation (group size=128) quantization for 4-bit quantization (denoted as W4A4).
We evaluate the perplexity
and the zero-shot accuracy on six tasks:
LAMBADA~\cite{radford2019language}, 
HellaSwag~\cite{zellers2019hellaswag},
PIQA~\cite{bisk2020piqa},
Arc (Easy and Challenge)~\cite{clark2018think},
Winogrande~\cite{sakaguchi2021winogrande},
and OpenbookQA~\cite{mihaylov2018can}
using lm-eval-harness~\cite{eval-harness}.

\begin{figure}[t]
    \centering    
    \includegraphics[width=0.7\columnwidth]{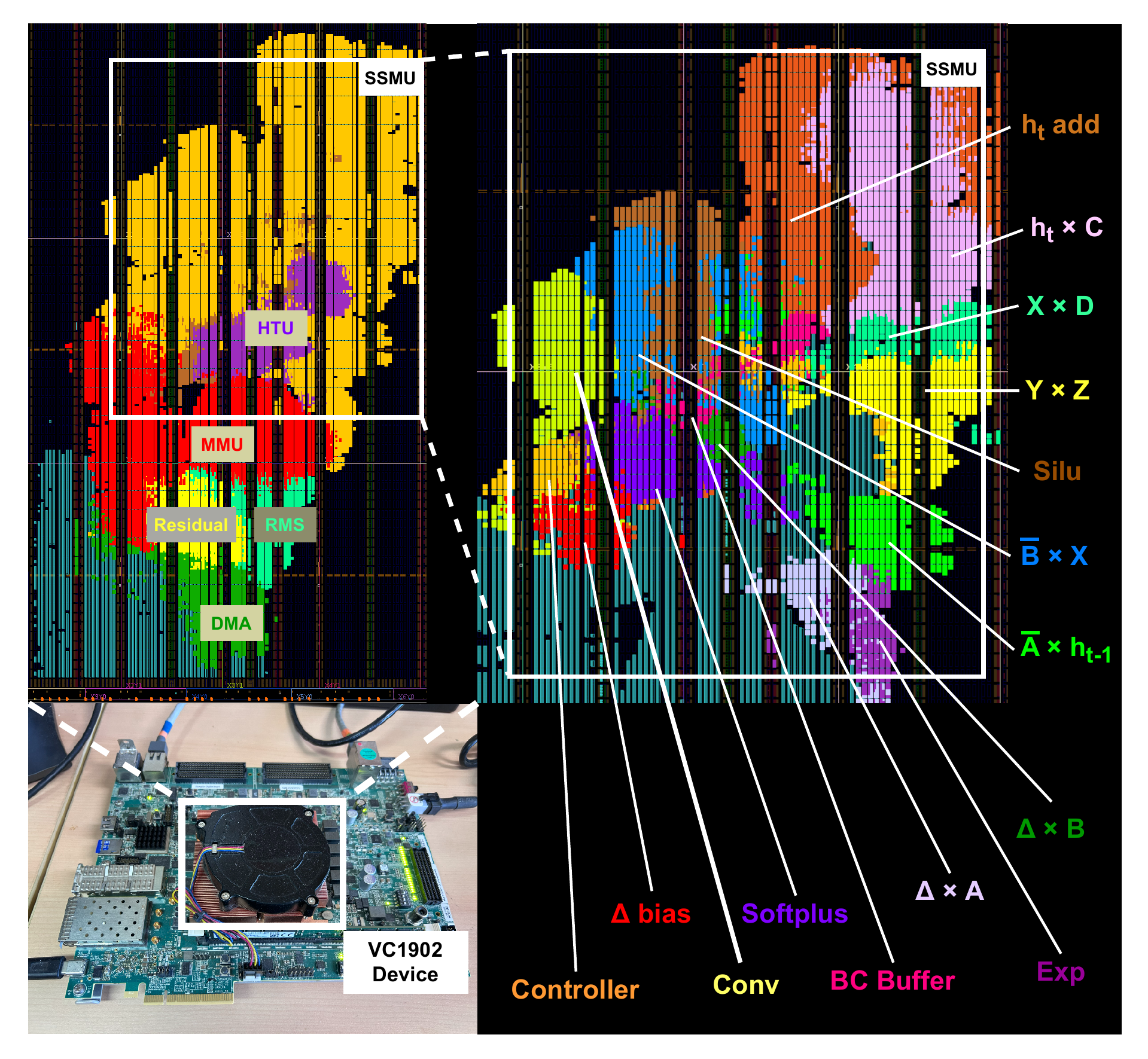}
    \caption{LightMamba implementation layout on VCK190.}
    \vspace{-10pt}
    \label{fig:FPGA_layout}
\end{figure}

\textbf{Hardware}
We use two FPGA platforms for evaluation: Xilinx Versal VCK190 and Alveo U280. 
We implement LightMamba on VCK190 using Vitis HLS
and Vivado Design FLow.
Fig.~\ref{fig:FPGA_layout} shows the layout
of our implementation on VCK190 FPGA.
We measure the throughput on-board using the PYNQ framework
and power consumption with the Xilinx BEAM tool.
For U280 evaluation, we develop a cycle-accurate simulator, 
which has been verified through HLS emulation using Vitis 2023.2.
We choose NVIDIA RTX 2070 and RTX 4090 as our GPU baselines.
We use NVIDIA system management interface for power 
measurements. 
Table ~\ref{tab:compare_with_prior art work} shows the hardware parameters of FPGA and GPU, as well as the detailed hardware utilization of LightMamba.

\subsection{Evaluation Result}
\label{subsec:Evaluation result}


\textbf{Algorithm Evaluation}
We compare our quantization algorithm 
with the prior art weight-activation PTQ methods
SmoothQuant (SQ)~\cite{xiao2023smoothquant}
and Outlier Suppression+ (OS+)~\cite{wei2023outlier} for Transformer-based LLM.
Note that these methods can only be applied to linear layers.
We re-implement them on Mamba
and use 128 random samples from WikiText2~\cite{merity2016pointer} dataset as the calibration data.
We evaluate our method LightMamba which quantizes only linear layers,
and LightMamba* which quantizes all modules including SSM.
As shown in Table~\ref{tab:acc_result}, for W8A8 quantization,
LightMamba and LightMamba* have negligible accuracy loss compared to the FP16 model.
Although OS+ has better perplexity on the Lambada dataset, it collapses on W4A4.
While our methods LightMamba and LightMamba* outperform all other methods on W4A4,
improving the perplexity by 1.78 and 1.91 compared to the prior art method SQ, respectively.

\begin{table*}[!tb]
    \centering
    \caption{Performance comparison of different methods on Mamba2-2.7B. 
    Only LightMamba* quantizes the entire model including SSM while others only quantize linear layers.
    The \textbf{bold} denotes the best and 
    the \underline{underlined} denotes the second-best.
    }
    \label{tab:acc_result}
    \scalebox{0.9}{
    \vspace{3pt}
    \begin{tabular}{c|c|c|c|c|c|c|c|c|c|c}
    \toprule
   \multirow{2}{*}{Method} & \multirow{2}{*}{Bit-precision} &LAMBADA &LAMBADA &HellaSwag & PIQA &Arc-E &Arc-C  &Winogrande  &OpenbookQA &Average \\
   &&ppl ↓&acc ↑&acc ↑&acc ↑&acc ↑&acc ↑&acc ↑ &acc ↑ &acc ↑\\
    \midrule
    FP16 & - &4.10 &69.7 &66.6 &76.4 &69.6 &36.4 &64.0 &38.8 &60.2 \\
      \midrule
      
     RTN &W8A8 &4.26 &68.8 &66.1 &75.8 &68.4 &36.4 &63.6 &38.4 &59.6\\
     SQ &W8A8 &4.28 &68.2 &66.0 &75.9 &69.1 &37.0 &63.4 &38.2 &59.7\\
     OS+ &W8A8 &\textbf{4.01} & 69.9 &66.2 &76.4 &69.5 &36.5 &63.4 &39.0 &\underline{60.1} \\
     LightMamba &W8A8 &4.07 &69.7 &66.5 &76.1 &69.3 &36.9 &64.0 &38.8 &\textbf{60.2}\\
     LightMamba* &W8A8 &\underline{4.03} &70.2 &66.2 &76.1 &69.4 &36.1 &64.6 &38.6 &\textbf{60.2} \\

    \midrule
     RTN &W4A4 &17.46 &37.7 &62.6 &70.1 &60.1 &34.5 &57.7 &38.2 &51.6 \\
     SQ &W4A4 &8.26 &53.4 &64.0 &73.6 &63.7 &35.1 &59.4 &39.0 &55.5 \\
     OS+ &W4A4 &$> 100$ & 0.0 &27.7 &54.5 &30.8 &24.7 &48.8 &25.6 &30.3 \\
     LightMamba &W4A4 &\underline{6.48} &57.3 &62.7 &73.5 &65.5 &35.3 &60.7 &37.6 &\textbf{56.3}\\
     LightMamba* &W4A4 &\textbf{6.35} &59.6 &62.4 &74.4 &64.7 &34.3 &59.9 &36.4 &\underline{55.9} \\
     
    \bottomrule
    \end{tabular}
    }
    \vspace{-10pt}
\end{table*}

\begin{figure}[t]
    \centering
   \vspace{-10pt} \includegraphics[width=1\columnwidth]{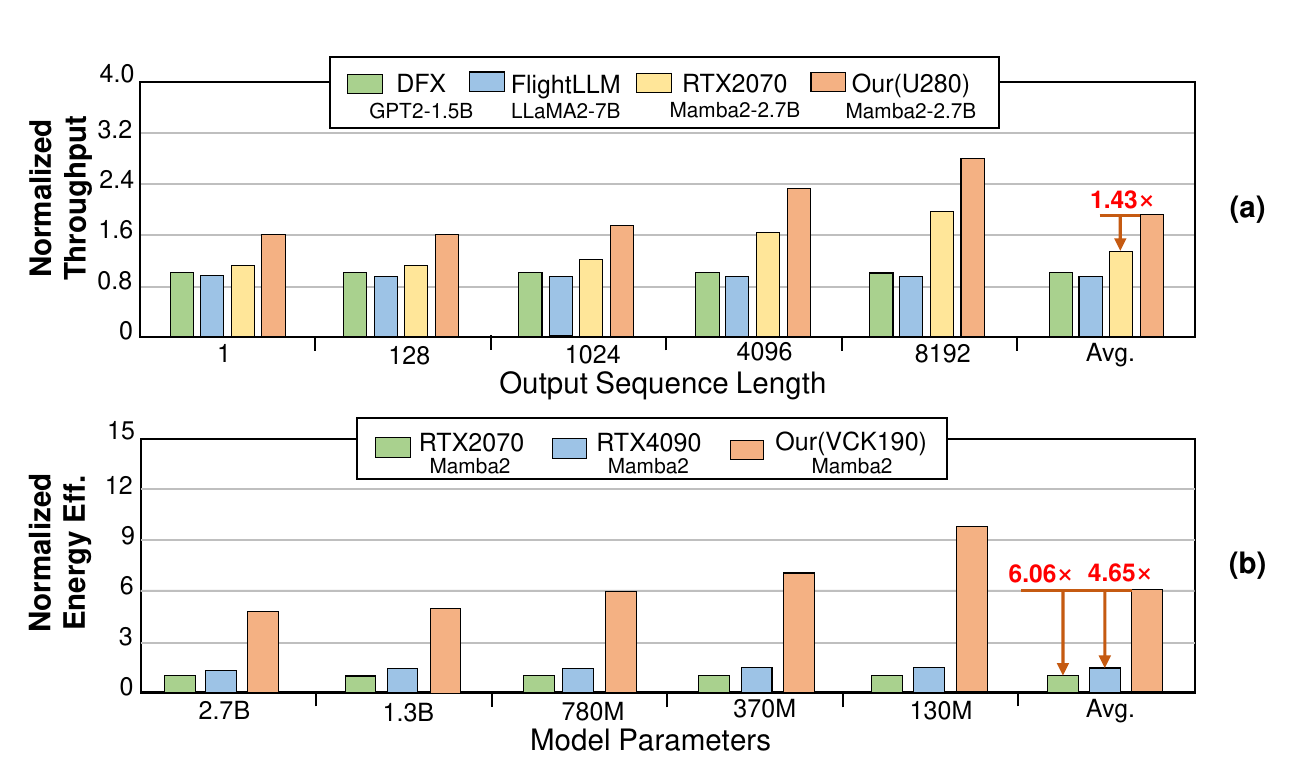}
    \caption{
    (a) Throughput with different output sequence length.
    (b) Energy efficiency with different model sizes.
    }
    \label{fig: model size}
\end{figure}

\textbf{Hardware Evaluation}
We compare the decoding throughput of GPUs, prior art accelerators, and our LightMamba on different output sequence lengths in Fig.~\ref{fig: model size}(a). 
As prior art accelerators have not supported Mamba,
we compare their performance when running Transformer-based LLMs.
Since these works did not provide throughput data for long sequence length,
we simulated their performance based on the parameters in each paper.
On VCK190, LightMamba achieves the practical throughput of 3.61 and 7.21 tokens/s for
W8A8 and W4A4, respectively.
We also simulate LightMamba on U280 for fair comparison. 
The throughput of LightMamba achieves \textbf{93} tokens/s, which outperforms RTX 2070 by $\textbf{1.43}\times$ on average. LightMamba achieves more significant acceleration on long sequences as Mamba only records hidden states of a fixed size.

We compare the energy efficiency (Tokens/J) of LightMamba and GPU on different model sizes in Fig.~\ref{fig: model size}(b).
LightMamba on VCK190 consistently outperforms GPUs and achieves on average $\textbf{6.06}\times$ and $\textbf{4.65}\times$ improvement over RTX 2070 and 4090 GPU, respectively.
For small Mamba models, LightMamba achieves more energy saving as our design reduces the overhead of the SSM layer,
which is more costly for small models.

\begin{table}[!t]
\centering
\caption{Hardware comparison with GPU.}
\label{tab:compare_with_prior art work}
\resizebox{0.45\textwidth}{!}{
\begin{tabular}{l|ccc|cc}
\toprule
& \multicolumn{3}{c|}{\textbf{LightMamba}} & \multicolumn{2}{c} {\textbf{GPU Baseline}}\\ 
\midrule
\textbf{Platforms}& VCK190 & VCK190  & U280  & RTX 2070 &RTX 4090\\
\midrule
\textbf{Frequency}    & 400MHz & 400MHz & 200MHz  & 1.62GHz & 2.52GHz \\
\textbf{Bandwidth}    & 12GB/s  & 12GB/s  & 460GB/s & 468GB/s & 1008GB/s     \\
\midrule
\textbf{Precision}   & W4A4   & W8A8  & W4A4  & FP16 & FP16\\

\textbf{LUT}          & 107k  & 111k  & 297k    & -   & -    \\
\textbf{FF}           & 130k   & 134k   & 394k    & -   & -     \\
\textbf{DSP}          & 228      & 228      & 1164     & -   & - \\
\textbf{BRAM}        & 912       & 914       & 912      & -   &  -    \\
\textbf{URAM}         & 61        & 61        & 61       & -    &   -   \\
\hline
\rowcolor{gray!20}
\textbf{Througput}    & 7.21     & 3.61      & \textbf{93}      & 65     & 138     \\

\rowcolor{gray!20}
\textbf{Energy Eff.}     & \textbf{2.25}     & \textbf{1.45}    & -    & 0.371  & 0.484     \\
\bottomrule
\end{tabular}
}
\vspace{-10pt}
\end{table}

\subsection{Ablation Study}
\label{subsec:Ablation Study}

\begin{figure}[t]
    \centering
    \includegraphics[width=\columnwidth]{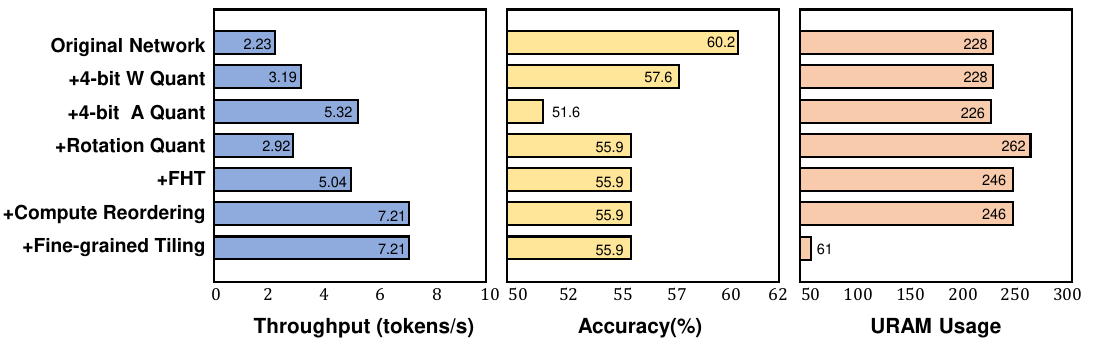}
    \caption{Impact of different techniques on the computation throughput, accuracy and URAM usage.}
    \vspace{-10pt}
    \label{fig:ablation}
\end{figure}

We now conduct the ablation study to show the accuracy and efficiency impact of different techniques in LightMamba.
As shown in Fig.~\ref{fig:ablation}, through weight and activation quantization, the throughput can be increased
from 2.23 to 5.32 tokens/s. With the rotation-assisted quantization algorithm and customized HTU, we
boost the accuracy of quantized Mamba by 4.3\% with almost the same throughput. 
Our computation re-ordering technique
improves the hardware utilization and raises the throughput further to 7.21.
Finally, through fine-grained tiling and fusion, the on-chip memory consumption is reduced significantly by 
4$\times$ from 246 to 61.

\section{Conclusion}


In this paper, we propose LightMamba,
an efficient FPGA-based Mamba acceleration framework
featuring quantization and hardware co-design.
We point out three challenges of accelerating Mamba on FPGA.
To solve them, we propose an FPGA-friendly rotation-assisted PTQ algorithm
quantizing Mamba to 4-bit with minimal accuracy degradation.
We further propose an FPGA accelerator with computation reordering 
and fine-grained tiling and fusion.
With these methods, LightMamba achieves 7.21 tokens/s on VCK190 FPGA and 4.65$\sim$6.06$\times$ higher energy efficiency over the GPU baseline.

\bibliographystyle{IEEEtran}
\bibliography{quantization,hardware}

\end{document}